%% file: main.tex
\documentclass[10pt,twocolumn,letterpaper]{article}

\usepackage{wacv}
\usepackage{times}
\usepackage{epsfig}
\usepackage{graphicx}
\usepackage{amsmath}
\usepackage{amssymb}

\wacvfinalcopy %

\usepackage{algorithm2e}
\usepackage{amsfonts}       %
\usepackage{booktabs}       
\usepackage{caption}
\usepackage[T1]{fontenc}    %
\usepackage[utf8]{inputenc} %
\usepackage{microtype}      %
\usepackage{multicol}
\usepackage{multirow}
\usepackage{nicefrac}       %
\usepackage{siunitx}
\usepackage{tabularx}
\usepackage{svg}
\usepackage{url}            %
\usepackage{xcolor}
\usepackage{xspace}

\usepackage{color, colortbl}
\definecolor{LBlue}{rgb}{.8,.8,1.}
\definecolor{MBlue}{rgb}{.6,.6,1.}
\definecolor{HBlue}{rgb}{.2,.2,1.}

\newcommand{\B}[1]{\textbf{#1}}

\newcommand{\ImNetA}{\textsc{ImageNet-A}\xspace}
\newcommand{\ImNetC}{\textsc{ImageNet-C}\xspace}
\newcommand{\ImNetClean}{\textsc{ImageNet}\xspace}
\newcommand{\ImNetVid}{\textsc{ImageNet-Vid}\xspace}
\newcommand{\ImNetVV}{\textsc{ImageNet-V2}\xspace}
\newcommand{\ImNetVidW}{\textsc{ImageNet-Vid-}\texttt{pm-k}\xspace}
\newcommand{\ObjetNet}{\textsc{ObjectNet}\xspace}
\newcommand{\YTeightM}{\textsc{YT8M}\xspace}  %
\newcommand{\YTBB}{\textsc{YT-BB-Robust}\xspace}
\newcommand{\YTBBW}{\textsc{YT-BB-Robust-}\texttt{pm-k}\xspace}
\newcommand{\YTBBSHORT}{\textsc{YT-BB}\xspace}
\newcommand{\YTBBWSHORT}{\textsc{YT-BB-}\texttt{pm-k}\xspace}

\newcommand{\resnet}{\textsc{ResNet50}\xspace}
\newcommand{\retinanet}{\textsc{RetinaNet}\xspace}
\newcommand{\openimages}{\textsc{OpenImagesV4}\xspace}

\usepackage[breaklinks=true,bookmarks=false]{hyperref}

\usepackage{cleveref}

\begin{document}
\title{Representation learning from videos in-the-wild: An object-centric approach\thanks{This work was published at IEEE WACV 2021}}

\author{%
  Rob Romijnders\thanks{Work done as part of Google AI residency.} \and Aravindh Mahendran \and Michael Tschannen\thanks{Work done while at Google Research. } \and Josip Djolonga \and Marvin Ritter \and Neil Houlsby \and Mario Lucic
}

\maketitle
\thispagestyle{empty}  %

\begin{abstract}
  We propose a method to learn image representations from uncurated videos. We combine a supervised loss from off-the-shelf object detectors and self-supervised losses which naturally arise from the video-shot-frame-object hierarchy present in each video. We report competitive results on 19 transfer learning tasks of the Visual Task Adaptation Benchmark (VTAB), and on 8 out-of-distribution-generalization tasks, and discuss the benefits and shortcomings of the proposed approach. In particular, it improves over the baseline on all 18/19 few-shot learning tasks and 8/8 out-of-distribution generalization tasks. Finally, we perform several ablation studies and analyze the impact of the pretrained object detector on the performance across this suite of tasks.
\end{abstract}

\section{Introduction}
\vspace{1mm}

Learning transferable visual representations is a key challenge in computer vision. The aim is to learn a representation function once, such that it may be transferred to a plethora of downstream tasks. In the context of image classification, models trained on large amounts of labeled data excel in transfer learning~\cite{bigtransfer}, but there is a growing concern that this approach may not be effective for more challenging downstream tasks~\cite{DBLP:journals/corr/abs-1910-04867}. Recent advances, such as  contrastive self-supervised learning combined with strong data augmentation, present a promising avenue~\cite{DBLP:journals/corr/abs-2003-04297}. 

\begin{figure}[t!]
    \centering
    \includegraphics[width=0.9\linewidth]{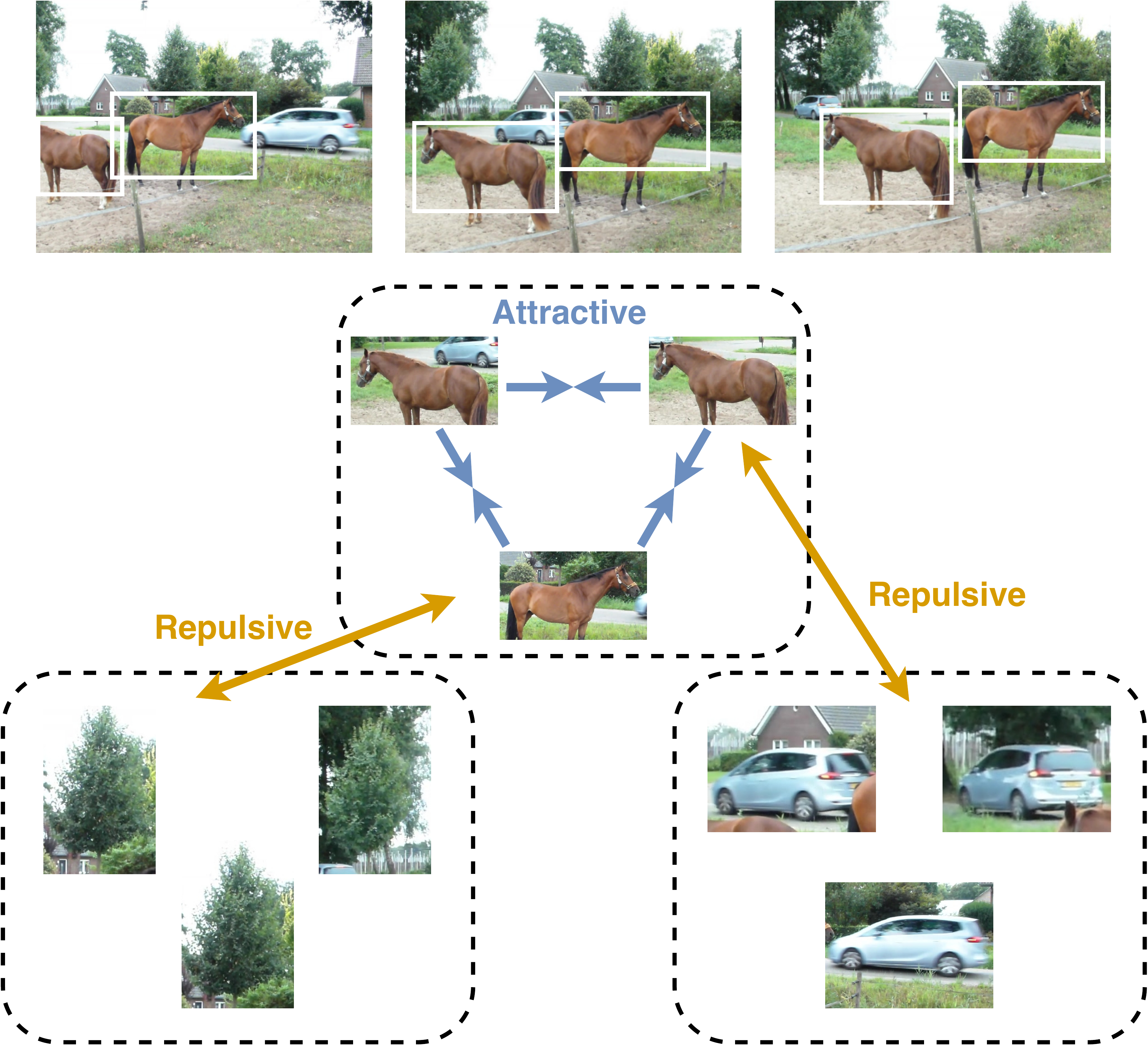} 
    \caption[Caption for LOF]{We propose to learn from objects in video. An off-the-shelf pretrained object detector tags each frame with bounding boxes and class labels. A contrastive loss encourages pairs of objects with the same class label (\emph{positives}) to be embedded closer to each other than those from disparate ones (\emph{negatives}). We augment existing work in video representation learning~\cite{vivi} with an object level loss.\looseness=-1  \label{fig:bbox_loss}}\vspace{-2mm}
\end{figure}

We consider the problem of learning image representations from uncurated videos. While these videos are noisy and unlabeled, they contain abundant natural variations of the objects present. Furthermore, videos decompose temporally into a hierarchy of videos, shots, and frames, which can be used to define pretext tasks for self-supervised learning~\cite{vivi}. We extend this hierarchy to its natural continuation, namely, the spatial decomposition of frames into objects. We then use the ``video, shot, frame, object'' hierarchy to define a more holistic pre-text task. In this setting we are hence given uncurated videos and an off-the-shelf pre-trained object detector, and we propose a method of supplementing the loss function with cues at the object level.

Videos, at the frame and shot level, convey global scene structure, and different frames and shots provide a \emph{natural data augmentation} of the scene. This makes videos a good fit for contrastive learning losses that rely on heavy data augmentation for learning scene representations~\cite{DBLP:journals/corr/abs-2002-05709}. At the object level, videos also provide rich information about the structure of individual objects. This can be valuable for tasks such as orientation estimation, counting, and object detection. Furthermore, object-centric representations can generalize to scenes constituted as a novel combination of known objects. Intuitively, each occurrence of the object forms a natural augmentation for objects of that class. Finally, one can make use of the fact that the same object appears in consecutive frames to enable representations which are more robust to perturbations and distributions shifts. Contrastive learning in this setting is illustrated in \Cref{fig:bbox_loss}. 

\vspace{-3mm}
\paragraph{Our contributions:}
\begin{enumerate}
    \item We extend the framework from VIVI~\cite{vivi} to include object level cues using an off-the-shelf object detector.
    \item We demonstrate improvements using object level cues on recent few-shot transfer learning benchmarks and out-of-distribution generalization benchmarks. In particular, the method improves over the baseline on all 18/19 few-shot learning tasks and 8/8 out-of-distribution generalization tasks.
    \item We ablate various aspects of the setup to reveal the source of the observed benefits, including (i) randomizing the object classes and locations, (ii) using only the object labels, (iii) using the detector as a classifier in a semi-supervised setup, (iv) cotraining with \ImNetClean labels, and (v) using larger ResNet models.
\end{enumerate}

\vspace{1mm}
\section{Related work} \label{sec:relatedwork}
\vspace{2mm}

\paragraph{Self-supervised image representation learning.}
The self-supervised signal is provided through a pretext task (e.g.\ converting the problem to a supervised problem), such as reconstructing the input~\cite{hinton2006reducing}, predicting the spatial context ~\cite{DBLP:conf/iccv/DoerschGE15, DBLP:conf/iccv/NorooziPF17}, learning to colorize the image~\cite{DBLP:conf/eccv/ZhangIE16}, or predict the rotation of the image~\cite{DBLP:conf/iclr/GidarisSK18}. Other popular approaches include clustering~\cite{DBLP:conf/eccv/CaronBJD18, DBLP:conf/iccv/ZhuangZY19, DBLP:conf/nips/DosovitskiySRB14} and recently generative modeling~\cite{DBLP:conf/iclr/DonahueKD17, DBLP:conf/nips/KingmaMRW14, bigbigan}. A promising recent line of work casts the problem as mutual information maximization of representations of different views of the same image~\cite{DBLP:journals/corr/abs-1807-03748}. These views can come from augmentations or corruptions of the same input image~\cite{DBLP:conf/iclr/HjelmFLGBTB19,DBLP:conf/nips/BachmanHB19,DBLP:journals/corr/abs-2002-05709,DBLP:conf/cvpr/He0WXG20}, or by considering different color channels as separate views~\cite{DBLP:journals/corr/abs-1906-05849}. 

\vspace{-3mm}
\paragraph{Representation learning from videos.}
The order of frames in a video provides a useful learning signal ~\cite{DBLP:conf/eccv/MisraZH16,DBLP:conf/iccv/LeeHS017,DBLP:conf/cvpr/FernandoBGG17,DBLP:conf/cvpr/WeiLZF18}. Information from temporal ordering can be combined with spatial ordering to infer object relations ~\cite{DBLP:conf/iccv/0004HG17} or co-occurrence statistics ~\cite{DBLP:journals/corr/IsolaZKA15}. Other pretext tasks include predicting the playback rate~\cite{yao2020video}, or clustering~\cite{DBLP:conf/cvpr/LeeLNN20}.

Orthogonal to the pretext tasks, one could use the paradigm of \textit{slow feature learning} in videos. This line of work dates back to \cite{DBLP:journals/neco/WiskottS02}, which developed a method to learn slow varying signals in time series, and inspired several recent works~\cite{DBLP:conf/iccvw/HanXZ19,DBLP:conf/cvpr/JayaramanG16,DBLP:conf/nips/ZouNZY12,DBLP:journals/corr/abs-2003-07990}. Our loss at the frame level uses insights from \textit{slow feature learning} in the form of temporal coherence~\cite{DBLP:conf/icml/MobahiCW09, DBLP:conf/iccv/RamanathanTML15}. \looseness=-1

Tracking patches is an alternative form of supervision which has some similarity with our object level loss. For example, \cite{DBLP:conf/iccv/WangG15,DBLP:conf/iccv/0004HG17,DBLP:conf/accv/GaoJG16} learn temporally coherent representations for the patches. Our method learns representations for the objects within a fully convolutional neural network. Other approaches have investigated learning specific structures to represent the objects in video ~\cite{DBLP:conf/nips/MindererSVCML19,DBLP:journals/corr/abs-2003-01460}.

Predicting the next frame or learning to synthesize a (future) frame were also considered as pretext tasks~\cite{DBLP:journals/corr/GoroshinBTEL15, DBLP:conf/icml/SrivastavaMS15,DBLP:journals/corr/MathieuCL15}. Given that  frame prediction requires learning of fine-detailed features, one can predict only the moving pixels~\cite{yao2020video}, or turn to time-agnostic prediction~\cite{DBLP:conf/iclr/JayaramanEEL19}.

\vspace{-3mm}
\paragraph{Object level supervision.} In terms of self-supervision we follow~\cite{vivi} to learn from the natural hierarchy present in the videos and make use of the losses studied therein. In contrast to~\cite{vivi} we incorporate object-level information in the final loss and show that it leads to benefits both for few-shot transfer learning and out-of-distribution generalization. Incorporating the pixel, object, and patch information for learning and improving \emph{video representations} was also considered in ~\cite{DBLP:conf/cvpr/WangJE19,DBLP:conf/iccv/WangG15,DBLP:conf/nips/ZouNZY12,zou2011unsupervised,DBLP:conf/accv/GaoJG16}. In contrast to these works, we do not rely on a strong tracker trained on a similar distribution, but on an off-the-shelf, parameter efficient object detector. Furthermore, we learn representations for images, not videos. Contemporary works also use object information for learning video representations~\cite{DBLP:conf/bmvc/LaiX19} or for training graph neural networks on videos~\cite{DBLP:conf/eccv/WangG18}.
\begin{figure*}[t!]
    \centering
    \includegraphics[width=\linewidth]{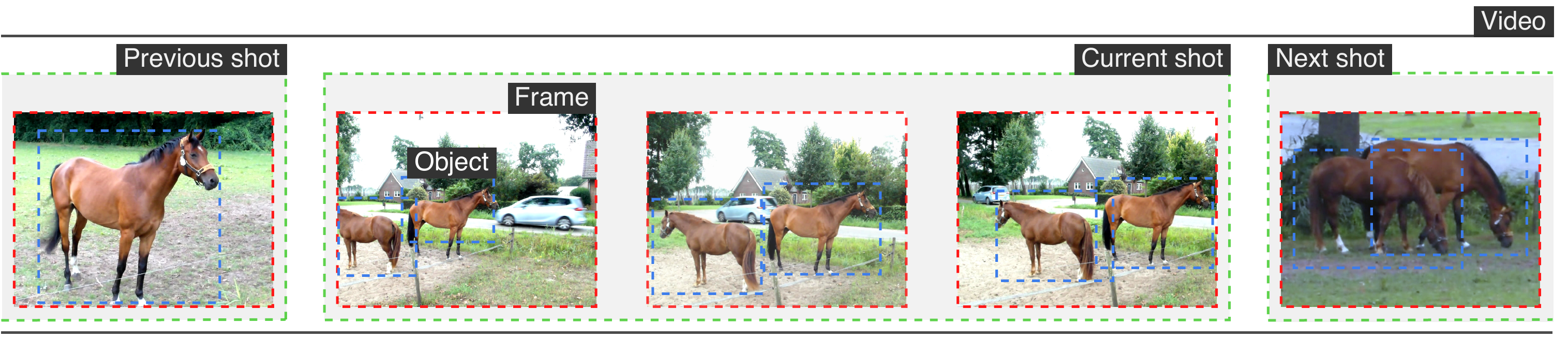}
    \caption{Learning from the natural hierarchy present in the videos. Each video in a dataset consists of multiple shots (indicated in the gray boxes), each shot consists of multiple frames. This hierarchy can be used to formulate a contrastive loss for learning image representations~\cite{vivi} (cf. Section~\ref{sec:method}). We extend this hierarchy to the object level by using an off-the-shelf detector.}
    \label{fig:learning_hieararchy}
\end{figure*}

\vspace{2mm}
\section{Method} \label{sec:method}
\vspace{2mm}

\paragraph{Self-supervision via video-shot-frame hierarchy.} A video can be decomposed into a hierarchy of shots, frames and objects which is illustrated in~\Cref{fig:learning_hieararchy}. For the first two levels in the hierarchy, we follow the setup from~\cite{vivi}, named \emph{VIVI}, which we summarize here. 

At the \textbf{shot level}, \textit{VIVI} learns, in a contrastive manner, to embed shots such that they are predictive of other shots in the same video \cite{DBLP:journals/corr/abs-1807-03748}. At the frame level, \textit{VIVI} learns to embed frames such that frames from the same shot are closer to each other relative to frames from other shots. Following the findings in~\cite{vivi}, the shot level loss is an instance of the InfoNCE loss~\cite{DBLP:journals/corr/abs-1807-03748} between shot representations. \textit{VIVI} (1) maps frame $k$ in shot $\ell$ to a representation $f_{k,\ell}^i$ in video $i$, (2) aggregates these frame representations into a shot representation $s_\ell^i$, and (3) predicts the representation of the next shot, $\hat{s}_{\ell+1}^i$, given the sequence of preceding shots, $s_{1:\ell}^i$, resulting in the loss:
\begin{align}\label{eq:shot_loss}
     \mathcal{L}_{\text{shot}} = -\frac{1}{N}\sum_{i,\ell,m} \log \frac{e^{g(\hat{s}_{\ell+m}^i, s_{\ell+m}^i)}}{\frac{1}{N} \sum_j e^{g(\hat{s}_{\ell+m}^i, s_{\ell+m}^j)}},
\end{align} 
where $g(\cdot, \cdot)$, called the critic, is used to compute similarity between shots, $N$ indicates the total numbers of videos in a mini batch, and $m$ indicates the number of prediction steps into the future. In practice, optimization is more stable when contrasting against shot representations from the entire batch of videos. \looseness=-1

Contrastive learning is also applied at the~\textbf{frame level} based on the intuition that frames within a shot typically contain similar scenes. \textit{VIVI} learns to assign a higher similarity to frame representations coming from the same shot by applying a triplet loss defined in \cite{DBLP:conf/cvpr/SchroffKP15} (cf. Figure~\ref{fig:learning_hieararchy}). In particular, for a fixed frame, frames coming from the same shot are considered as positives, while the frames from other shots are negatives. Denoting positive pairs as $f_{k, \ell}^{p}$ and negatives as $f_{k, \ell}^n$, the semi-hard loss can be written as~\cite{DBLP:conf/cvpr/SchroffKP15}:  
\begin{equation*}%
         \mathcal{L}_{\text{frame}} = \sum_{k, \ell} \max\{\left\|e_{k, \ell}-e_{k, \ell}^p\right\|_2^2 -
                  \left\|e_{k, \ell}-e_{k\ell}^n\right\|_2^2+\alpha, 0 \}\;
\end{equation*}

\paragraph{Extending the hierarchy with object-level losses.}\label{sec:bbox_loss}
Data augmentation is a key novelty behind recent advances in representation learning~\cite{DBLP:journals/corr/abs-1912-01991,DBLP:journals/corr/abs-2002-05709,DBLP:journals/corr/abs-1906-05849,autoaugment}. These augmentations are usually obtained by applying synthetic transformations like random cropping, left-right flipping, color distortion, blurring, or adding noise. However, independent non-rigid movement of an object against background, as seen in real video data, is hard to expect from synthetic augmentations. 

In order to exploit these natural augmentations, which occur in video, we use a contrastive loss that encourages representations of objects of the same category to be closer together as opposed to representations of different categories (cf. Figure~\ref{fig:bbox_loss}). To construct this loss, we apply an off-the-shelf object detector to all frames and extract the bounding boxes and class labels. Given the representations of each bounding box (will be discussed later), we use a triplet loss where objects from the same class form positive pairs, and objects from different classes form negative pairs. 
In particular, given the embedding of the $b$-th bounding box $r_b$, and the embeddings of the positive $r_p$ and negative $r_n$ (with respect to $r_b$), we apply the following loss for each frame:
\begin{equation}\label{eq:object_loss}
 \mathcal{L}_{\textsc{object}} = \sum_b \max \{ \left\|r_b-r_{p}\right\|_2^2 -
                  \left\|r_b-r_{n}\right\|_2^2+\alpha, 0\}\;
\end{equation}
To scale the triplet loss to a large number of boxes, we follow insights from the literature ~\cite{zou2011unsupervised,DBLP:conf/cvpr/KolesnikovZB19,DBLP:conf/cvpr/SongXJS16} and use semi-hard negative mining ~\cite{DBLP:conf/cvpr/SchroffKP15}. \looseness=-1 

An alternative to the contrastive loss is the classic cross-entropy loss --- learning a representation such that a linear classifier can recognize the target class. We formalize the former given the empirical evidence from recent research~\cite{DBLP:conf/nips/KhoslaTWSTIMLK20}, but we present ablation studies in Section~\ref{sec:experiments}.

\begin{figure}[b]
    \centering
    \includegraphics[width=\linewidth]{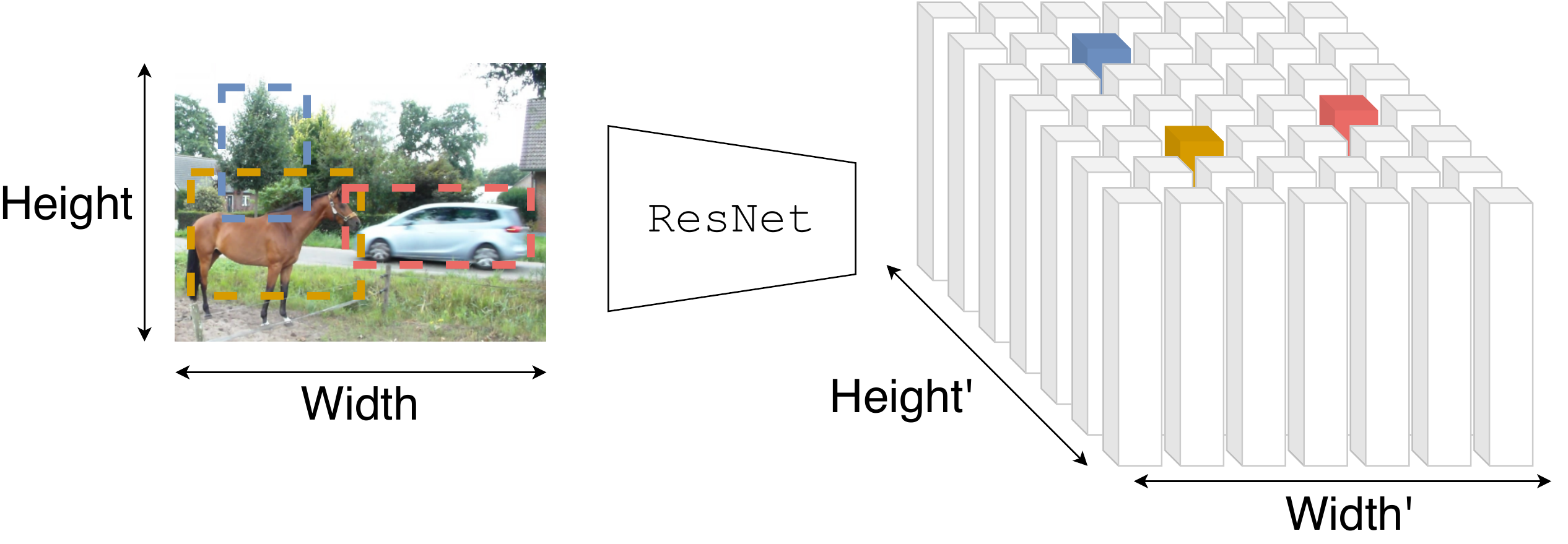}
    \caption{Illustration of the feature grid output by a \resnet. Object bounding boxes are mapped to feature columns that correspond to the center pixel of that bounding box. Correspondence is illustrated using matching colors.}
    \label{fig:feature_grid}
\end{figure}

\paragraph{Representations of bounding boxes.} A simple approach to obtain the representation would be to extract all the bounding boxes and feed them through the network. However, this is computationally prohibitive, and we instead propose a method which reuses the feature grid present in \resnet models~\cite{DBLP:conf/eccv/HeZRS16} illustrated in Figure~\ref{fig:feature_grid}.

Consider an image $x$ of dimensions $H \times W \times C$, indicating the height, width, and number of channels of the image, respectively. A fully convolutional \resnet maps this image to a feature grid $x'$ of dimensions $H' \times W' \times C'$. We represent the bounding box with center index $(i, j)$ by the vector $r = x'\left[ \lfloor i \frac{H'}{H} \rfloor, \lfloor j \frac{W'}{W} \rfloor \right]$ of size $C'$. This approach is conceptually similar to max pooling as used in Fast-RCNN~\cite{DBLP:conf/iccv/Girshick15}, and reminisces of \cite{DBLP:conf/eccv/ZhouKK20}. Given the computational efficiency and the fact that the effective receptive field is concentrated at the center~\cite{DBLP:conf/nips/LuoLUZ16}, we chose this simple alternative. 

\paragraph{Final loss function.} We combine the losses using positive coefficients $\omega$ and $\beta$ as
\begin{equation}\label{eqn:losses_combi}
    \mathcal{L} = \omega \mathcal{L}_{\textsc{object}} + \mathcal{L}_{\textsc{frame}} + \beta \mathcal{L}_{\textsc{shot}}.
\end{equation}
This formulation enables a study of the benefits of each of the losses and leads to practical recommendations.

\paragraph{Headroom analysis using \ImNetClean.}
In addition to the proposed method, we analyze the benefits of co-training the proposed network with a large labeled data source~\cite{vivi,DBLP:conf/iccv/BeyerZOK19}. This data source provides a vast quantity of labeled images and should help the model improve the performance on tasks which require fine-grained detail of specific object classes. In particular, we consider an affine map of the representation extracted by the network,  followed by a softmax layer and a corresponding cross-entropy resulting in
$\mathcal{L_{\textsc{total}}} = \mathcal{L} + \gamma \mathcal{L}_{\textsc{supervised}}$,
where $\gamma$ is a hyperparameter balancing the impact of this additional loss.

\vspace{2mm}
\section{Experimental setup} \label{sec:experiments}
\vspace{2mm}

\subsection{Architectures and training details} \label{sec:expdetails}

Unless otherwise specified, all experiments are performed on a \resnet V2~\cite{DBLP:conf/eccv/HeZRS16} with batch normalization. For the shot prediction function, we use a LSTM with 256 hidden units. We parameterize the critic function $g$ as a bilinear form. All frames are augmented using the same policy as \cite{DBLP:conf/cvpr/SzegedyLJSRAEVR15}, using random cropping, left-right flipping and color distortions. The coordinates of object bounding boxes are recalculated accordingly. All models are trained using a batch size of 512 for 120K iterations of stochastic gradient descent with a momentum constant of \num{0.9}. The learning rate starts as \num{0.8} and decreases by a factor of 10 after 90k and 110k training steps. When cotraining, we train for 100k iterations and decrease the learning rate after 70k, 85k and 95k iterations. Shots and frames are sampled using the same method as \cite{vivi}: for each video, we sample a sequence of four shots, and we sample eight frames from each shot.

The coefficients $\omega$ and $\beta$ weigh the loss contributions in~\Cref{eqn:losses_combi}. We set $\beta=0.04$ following \cite{vivi}, and $\omega=5$, although we found that a wider range of values leads to the same performance (cf.~\Cref{fig:sensitivity_weight} in the Appendix).

\paragraph{Cotraining details.} The experiments on cotraining use group normalization~\cite{DBLP:journals/ijcv/WuH20} with weight standardization~\cite{DBLP:journals/corr/abs-1903-10520}, instead of batch normalization, for a fair comparison to~\cite{vivi}. When cotraining, we sample at every step a batch from each dataset --- we compute the three-level loss \eqref{eqn:losses_combi} on the sampled videos, and the classification log-loss on the sampled \ImNetClean images. Cotrained models train with batch size of 512 for videos and 2048 for images for 100k iterations, using the learning rate schedule described above. Images are preprocessed using the \emph{inception crop} function from~\cite{DBLP:conf/cvpr/SzegedyLJSRAEVR15}.

\paragraph{Datasets.} We train on videos from the \YTeightM dataset and cotrain with \ImNetClean~\cite{imagenet}.
The videos are sampled at \num{1}Hz and we run the detector, a MobileNet~\cite{DBLP:conf/cvpr/SandlerHZZC18}, with a single shot multi box detector~\cite{DBLP:conf/eccv/LiuAESRFB16}, trained on \openimages~\cite{tfhub}. The detector runs at \num{19}ms per frame on a V100 GPU. We detected the objects offline and stored the annotated videos to disk for use during training. \Cref{tab:object_recurrence} in the appendix shows how often common objects are detected in the video frames. As the detector has been trained on \openimages, we use its 600 category label space for constructing positive and negative pairs for $\mathcal{L}_{\textsc{object}}$. We use the feature grid from the ResNet block 4 to construct representations for objects in a frame and limit the number of objects in each frame to a maximum of \num{5}. We discard objects with detection score below \num{0.05}, which accounts for approximately \num{3}\% of the detected objects. \Cref{fig:hist_det_score} shows a histogram of the detection scores. Finally, given that the \YTeightM dataset is a dynamic dataset, our training set contains the videos still available in May 2020: \num{3.3} million training and one million validation videos. The baselines were re-trained on this new dataset.
\input{table_vtab}

\subsection{Evaluation}
We evaluate two aspects of the learned representations: How well the representations transfer to novel classification tasks, and how robust are the resulting classifiers to distribution shifts.

\paragraph{Transferability.} The main objective of this work is learning image representations that transfer well to novel, previously unseen tasks. To empirically validate our approach, we report the results for transfer learning on the Visual Task Adaptation Benchmark (\textsc{VTAB}), a suite of 19 image classification tasks~\cite{DBLP:journals/corr/abs-1910-04867}. The tasks are organized into three categories, \emph{Natural} containing commonly used classification datasets (Caltech101, Cifar-100, DTD, Flowers102, Pets, SUN397 and SVHN), \emph{Specialized} comprising of images recorded with specialized equipment (Resisc45, EuroSAT, Patch Camelyon, Diabetic Retinopathy), and \emph{Structured} containing scene understanding tasks (CLEVR-dist, CLEVR-count, dSPRITES-orient, dSPRITES-pos, sNORB-azimuth, sNORB-elevation, DMLab, KITTI). For more details and references to the respective datasets, please refer to \cite{DBLP:journals/corr/abs-1910-04867}.

We consider transfer learning in the low data regime, where each task has only 1000 labeled samples available.
The evaluation protocol is the same as in~\cite{vivi,bigtransfer,DBLP:journals/corr/abs-1910-04867,vtabimplementation}: for each dataset we (i) train on \num{800} training samples using our learned model as initialization, (ii) sweep over two learning rates (\num{0.1}, \num{0.01}) and two learning rate schedules (10K steps with decay every 3K, or 2.5K steps with decay every 750), (iii) pick the learning rate and learning rate schedule according to the highest validation accuracy on the \num{200} validation samples, and then (iv) retrain the model using all \num{1000} samples. We report statistical significance at the $p=0.05$ level on a Welch's two sided $t$-test based on 12 independent runs of the transfer protocol. The error bars in the diagrams indicate bootstrapped 95\% confidence intervals.

\vspace{-2mm}
\paragraph{Robustness.}
As discussed in~\Cref{sec:method}, we were guided by the intuition that the model should learn to be more invariant to natural augmentations. We thus expect our model to be more robust and generalize better to out-of-distribution (OOD) images. 

We follow two recent studies on OOD generalization~\cite{DBLP:journals/corr/abs-1807-01697,bigrobustness} and evaluate robustness as accuracy on a suite of 8 datasets measuring various robustness aspects. These datasets are defined on the \ImNetClean label space: (1) \ImNetA~\cite{DBLP:journals/corr/abs-1907-07174} measures the accuracy on samples from the web that were adversarial to a \resnet trained from scratch on \ImNetClean. (2) \ImNetC~\cite{DBLP:journals/corr/abs-1807-01697} measures the accuracy on samples from \ImNetClean under perturbations such as blur, pixelation, and compression artifacts. (3) \ImNetVV~\cite{DBLP:conf/icml/RechtRSS19} presents a new test set for the \ImNetClean dataset. (4) \ObjetNet~\cite{DBLP:conf/nips/BarbuMALWGTK19} consists of images collected by crowd sourcing, where participants were asked to photograph objects in unusual poses and unusual backgrounds. (5-8) \ImNetVid, \ImNetVidW, \YTBB, and \YTBBW present frames from video sequences~\cite{DBLP:journals/corr/abs-1906-02168}. We measure both accuracy of the anchor frame, denoted as anchor accuracy, and worst-case accuracy in the 20 neighboring frames (i.e. if any frame from the set of 10 preceding and 10 following frames is misclassified, the video is judged as misclassified), denoted as \texttt{pm-k}. 

We also evaluate our models on the texture-shape data set from~\cite{DBLP:conf/iclr/GeirhosRMBWB19}. Our method uses a contrastive loss to learn specifically from objects. Learning with our loss encourages objects in different appearances to have similar representations. As such, we hypothesize that our models have higher shape bias, compared to texture bias. \cite{DBLP:conf/iclr/GeirhosRMBWB19} provide a dataset to measure the texture-shape bias. The test set consists of \num{1280} images whose texture has been stylized. Each image has a label according to its shape, and a label according to the stylization of its texture. We report the fraction of correct predictions based on shape, as proposed by the authors. For further details we refer to the paper~\cite{DBLP:conf/iclr/GeirhosRMBWB19}.

\vspace{2mm}
\section{Results} \label{sec:results}
\vspace{2mm}

\subsection{Transferability} 
\begin{figure*}[!t]
    \centering
    \includegraphics[width=\linewidth]{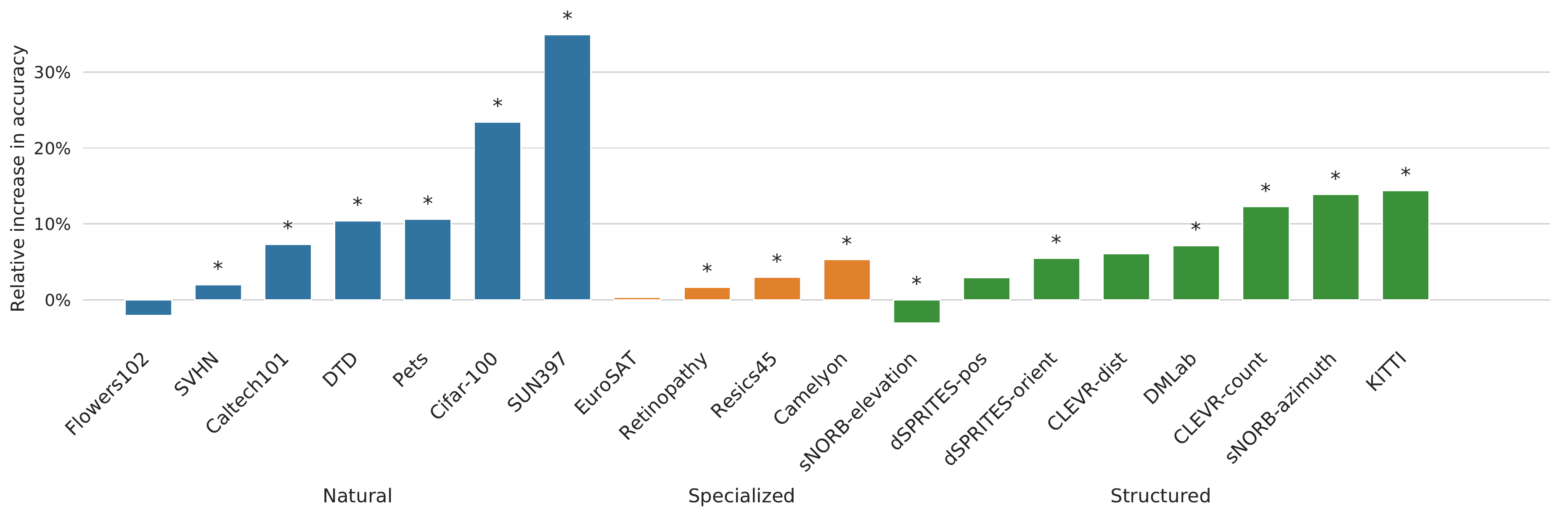}
    \caption{Relative increase in VTAB accuracy per dataset: blue for Natural, orange for Specialized and green for Structured datasets. Stars indicate statistical significance. Relative increase refers to the increase in accuracy by learning from objects, divided by accuracy of VIVI~\cite{vivi}, which only learns at two levels of the hierarchy.}
    \label{fig:relative_increase_per_dataset}
\end{figure*}

\Cref{tab:vtab} shows our results on the Visual Task Adaptation Benchmark (VTAB). We observe statistically significant increases in accuracy over the baseline~\cite{vivi} which demonstrate the benefits of supplementing the self-supervised hierarchy with object level supervision. The results per dataset are presented in~\Cref{fig:relative_increase_per_dataset}.\looseness=-1

Rows 1 and 2 in \Cref{tab:vtab} compare against two prior works on representation learning from videos: Transitive Invariance (TI) \cite{DBLP:conf/iccv/0004HG17} and Multi-task Self-Supervised Visual Learning (MT) \cite{DBLP:conf/iccv/DoerschZ17}. TI uses context based self-supervision together with tracking in videos to formulate a pretext task for representation learning and row 1 shows the performance of their pre-trained VGG-16 checkpoint. MT uses a variety of pretext tasks, including motion segmentation, coloring and exemplar learning~\cite{DBLP:journals/pami/DosovitskiyFSRB16} and row 2 shows the performance of their \textsc{ResNet101} (up to block 3) checkpoint.

\vspace{-2mm}
\paragraph{Ablation 1: Randomizing the location and the class.}
The object level loss $\mathcal{L}_\textsc{object}$ is made possible through additional supervision provided via an object detector pre-trained on \openimages. The detector contributes to representation learning by annotating object positions and object category labels in video frame and here we ablate these two sources: (i) We evaluate the contribution (1) from knowing the class of an object, but not its coordinates, and (2) when neither the class nor the location are known.

The results are detailed in~\Cref{tab:vtab}. Randomizing both the label and the coordinates of the objects destroys all signal from the detector. Row \textit{Boxes and labels at random} shows the results of this ablation and we observe that the performance is below the VIVI baseline, as expected. In contrast, when we randomize the object locations, but maintain the correct labels, we obtain an improvement over the baseline (row \textit{boxes at random}). Interestingly, the VTAB score on structured datasets, \num{59.7}\%, equals the accuracy where both the class and location are known.
\vspace{-2mm}
\paragraph{Ablation 2: Frame-level labels from a \ImNetClean-pretrained model.}
We further investigate the effectiveness of knowing frame-level labels by obtaining soft-labels using an \ImNetClean-pretrained model, effectively distilling the \ImNetClean model on \YTeightM frames~\cite{DBLP:journals/corr/HintonVD15}. Its performance is noted in~\Cref{tab:vtab}, row \textit{distilling from \ImNetClean}.
Interestingly, this distilled model scores higher in natural datasets, but lower in structured datasets than the proposed method. These differences show how various upstream signals affect different downstream tasks differently.

\begin{table}[t]
    \addtolength{\tabcolsep}{3pt}  %
    \centering
     \begin{tabularx}{\linewidth}{X l l l l} 
     \toprule 
     Method & \num{0}    & \num{1}k   & \num{20}k   & \num{120}k \\
     \midrule 
     VIVI   & 39.1  & 49.0 $^{\dagger}$ & 58.9 $^{\dagger}$ & 59.6 $^{\dagger}$   \\
     OURS   & 39.8  & 54.5 $^{\dagger}$ & 63.4 $^{\dagger}$ & 66.0 $^{\dagger}$   \\   
     \bottomrule
     \end{tabularx}
    \caption{\label{tab:rep_nn} Fraction of objects whose nearest neighbor in representation space is an object with the same class label for increasing number of training steps. $^{\dagger}$ indicates a statistically significant difference at $p=0.001$ using Fisher's exact test for all objects in \num{50} batches of \num{8} videos. As training progresses, our method has significantly more objects with matching neighbors than the vanilla VIVI model.}
\end{table}

\vspace{-3mm}
\paragraph{Ablation 3: Distilling the object detector.} We distill a \resnet on \YTeightM where the training instances are cropped objects and the labels assigned by the object detector. The distilled \resnet achieves a score of 57.1\% VTAB score compared to \num{64.1}\% of the proposed method. At the same time, using a non-pretrained ResNet of the same capacity achieves \num{42.1}\% when trained on 1000 downstream labels. Hence, the detector clearly provides a strong training signal, but it can be exploited to a higher degree by coupling it with a self-supervised loss as in the proposed method.

\vspace{-2mm}
\paragraph{Ablation 4: Semi-supervised learning.} One can also utilize the tagged frames as labelled training data~\cite{DBLP:conf/nips/KhoslaTWSTIMLK20}. To this end, we use a linear classifier to classify the bounding box representations $r_b$ as one of 600 \openimages classes using a binary cross-entropy loss added to the loss in~\Cref{eqn:losses_combi}. This approach increases the VTAB score from \num{64.1}\% to \num{64.9}\%. We also investigated using this loss as a \emph{replacement} for $\mathcal{L}_\textsc{object}$ in~\Cref{eqn:losses_combi}. However, this performed worse, scoring \num{63.7}\%, which highlights the advantage of the contrastive formulation.

\setlength{\tabcolsep}{4pt}
\begin{table}[t]
  \centering
  \normalsize
  \begin{tabularx}{\linewidth}{X r}
  \toprule
  Method & mAP(\%) \\
  \midrule 
  Our method ($\mathcal{L}_{\textsc{object}} + \mathcal{L}_{\textsc{frame}} + \beta \mathcal{L}_{\textsc{shot}}$)        & 40.4 \\
  VIVI ($\mathcal{L}_{\textsc{frame}} + \beta \mathcal{L}_{\textsc{shot}}$)                    & 35.1 \\
  Only object level ($\mathcal{L}_{\textsc{object}}$) & 39.3 \\
  \bottomrule
  \end{tabularx}
  \caption{\retinanet performance on the MS-COCO dataset using various pre-trained backbones.}
  \label{tab:retinanet}
\end{table}
\setlength{\tabcolsep}{1.4pt}

\paragraph{Effect of the contrastive loss.} Lastly, we present a diagnostic for our training procedure at the object level. $\mathcal{L}_\textsc{object}$ is designed to embed representations for objects of the same class closer together. We verify whether this is indeed the case by measuring the fraction of nearest neighbors for each representation that belongs to the same category. \Cref{tab:rep_nn} shows the progression of this metric during training, in comparison to a VIVI model trained in tandem. Our method results in a significantly higher fraction, indicating that more nearest neighbors belong to the same class as the query object. This verifies that our loss function and training procedure achieve the desired outcome. 

\vspace{+2mm}
\paragraph{Evaluation on detection.} Our model learns from videos at the object level. It is natural to expect that a \resnet backbone pre-trained using our method will perform well when fine-tuned for downstream object detection. To this end, we fine-tune a \retinanet architecture~\cite{DBLP:conf/iccv/LinGGHD17} on the MS COCO object detection dataset~\cite{DBLP:conf/eccv/LinMBHPRDZ14}. Images are rescaled and randomly cropped to $640 \times 640$ during training. We train the model for 60 epochs with an initial learning rate of \num{0.08} and batch size 256.

Results are shown in~\Cref{tab:retinanet}. Pre-training using our method improves upon the VIVI baseline by \num{5.3}\% mAP points. Training on only the object level loss scores \num{1.1}\% mAP point lower compared to using all three levels of the hierarchy. These results suggest that the learned representations are indeed more object centric and that learning from all three levels combined yields representations more effective for downstream object detection.

\paragraph{Co-training with \ImNetClean. } \Cref{tab:vtab_cotraining} shows the resulting accuracies on VTAB when cotraining with \ImNetClean. Compared to the cotrained VIVI baseline, our method with its object-level loss increases the VTAB score from \num{69.0}\% to \num{69.4}\%. This increase in accuracy is modest in comparison to those in \Cref{tab:vtab}.
We argue that \ImNetClean is a clean curated dataset whereas \YTeightM is noisy. Adding cotraining with clean \ImNetClean improves the accuracy on natural datasets from \num{60.9}\% to \num{70.3}\%. It is not surprising that adding more noisy supervision, at the object level, does not give massive gains in this setting. We repeat the experiment with a higher capacity \resnet. Again we observe modest, but statistically significant, improvements over VIVI. The largest improvement is on the structured datasets, which increase from \num{62.2}\% to \num{63.0}\%. These experiments highlight an interesting dichotomy between natural and structured subsets of VTAB: learning with \ImNetClean yields improvements on natural datasets, while using the detector yields improvements for structured datasets.

\begin{table}[t]
    \addtolength{\tabcolsep}{5pt}  %
    \centering
     \begin{tabularx}{\linewidth}{l llll} 
     \toprule 
     \textsc{Method}\quad\quad & \textsc{VTAB} & \textsc{Natural} & \textsc{Spec.} & \textsc{Struc.} \\
     \midrule
     \multicolumn{5}{l}{\small{\textit{ResNet-50}}}                                           \\
     VIVI \cite{vivi}         & 69.0 $^{\dagger}$ & 70.3 $^{\dagger}$  & 82.7 $^{\dagger}$  &  60.9   \\
     OURS                     & 69.4 $^{\dagger}$ & 70.8 $^{\dagger}$  & 82.9 $^{\dagger}$  &  61.4   \\     
     \midrule
     \multicolumn{5}{l}{\small{\textit{3x wider ResNet-50}}}                                           \\
     VIVI \cite{vivi}         & 70.2 $^{\dagger}$ & 71.4 $^{\dagger}$  & 83.7              &  62.2 $^{\dagger}$ \\
     OURS                     & 70.5 $^{\dagger}$ & 71.6 $^{\dagger}$  & 83.6              &  63.0 $^{\dagger}$ \\
     \bottomrule
     \end{tabularx}
    \caption{\label{tab:vtab_cotraining} VTAB scores of the models cotrained on \YTeightM and \ImNetClean. The presented numbers are the average image classification accuracy of the fine-tuned models over the respective VTAB category. ${\dagger}$ notes a statistically significant difference between VIVI and our method. }
\end{table}

\subsection{Robustness}

\Cref{tab:robustness_eval} presents the classification accuracies on the eight robustness datasets. To get predictions in the \ImNetClean label space, we fine-tune our learned representation and report results in row \textit{fine tuning}. 

Our method compares favorably to the baseline on all datasets, which confirms the intuition that extending the video-shot-frame hierarchy to objects results in more robust image representations. The robustness results for the cotrained models are presented in~\Cref{tab:robustness_eval}, row \textit{cotraining}. As expected, the results improve across all datasets. The final two columns of \Cref{tab:robustness_eval} note the delta between anchor accuracy and \texttt{pm-k} accuracy. A lower delta indicates a more robust model, irrespective of the anchor accuracy. In three out of four cases our method scores a lower (better) delta. \looseness=-1
\input{table_robustness}

\vspace{\baselineskip}
We evaluate our models on the texture-shape data set from~\cite{DBLP:conf/iclr/GeirhosRMBWB19}. As the evaluation is done using the \ImNetClean label space, we use the same models that we evaluated on the robustness datasets. First, we evaluate the fine-tuned models. The VIVI model, using only the video-shot-frame hierarchy, scores \num{20.6} shape fraction on the provided dataset. Using our method to learn from the video-shot-frame-object hierarchy, the shape fraction increases to \num{24.0}. A higher shape fraction indicates a better model, as the network has higher relative accuracy according to the shape of the object depicted. Similarly, cotrained models improve from \num{25.8} to \num{28.5} when using our method to learn from objects in video. These results indicate a promising direction for future research. \looseness=-1

\newpage
\section{Discussion}
 
We presented a hierarchy, videos-shots-frames-objects, to learn representations from video at multiple levels of granularity. The learned representations transfer better to downstream image classification tasks and exhibit higher accuracy on out-of-distribution datasets. We identify three aspects for future research.
\vspace{-4mm}
\paragraph{A taxonomy for learning transferable representations.} Our results show that using different signals from videos benefits transfer learning to Natural, Specialized or Structured image classification tasks in a specific manner. We consider our work in a larger line of research that creates a taxonomy for learning methods and their effect on transfer learning to specific datasets, similar to~\cite{taskonomy}, which outlined a taxonomy of multi modal learning. To give examples: Using the noisy videos from \YTeightM mainly improves transferability to Specialized and Structured tasks on the VTAB benchmark. Using the clean images from \ImNetClean improves transferability to Natural tasks.  Our method, which receives implicit supervision from \openimages, shows highest improvement on Natural tasks. Thus using different sources of supervision improves the transferability to different tasks. We believe that understanding how different data and learning methods impact the performance on different data domains is a central research question in transfer learning, and that this work contributes towards this grand challenge by providing insight into the benefits of learning from uncurated video data.

\vspace{-3mm}
\paragraph{Learning about objects without labels.} Our method uses an off-the-shelf detector to identify the objects. As the detector was trained on labeled data, learning at the object level of the hierarchy uses implicit supervision. Contemporary literature focused on other self-supervised methods to improve learning from video. For example, one could derive signals about objects using optical flow or keypoint detection \cite{DBLP:conf/nips/JabriOE20,jakab2020self}. Combining these ideas in our paradigm of learning in the hierarchy might provide a useful research direction.

\vspace{-4mm}
\paragraph{Learning about entire videos instead of image representations.} Our method shows improved results concerning transfer learning and robustness of image models for single images. This improvement raises the question how these results will translate to video understanding. Recently, there has been interest in video recognition~\cite{DBLP:journals/corr/KayCSZHVVGBNSZ17} and video action localization~\cite{DBLP:conf/cvpr/ZhukovACFLS19}. We look forward to testing our learning methods on these tasks. 

\vspace{-4mm}
\paragraph{Improved robustness from learning about objects.} We have shown how our method results in more robust image classifiers. This observation suggests that learning about objects, invariant to other parts of the images, improves robustness. Several computer vision tasks concern objects. Therefore, we suggest that having object centered representation will contribute to developments in robustness.

\ifwacvfinal
\section*{Acknowledgements} We thank Justin Gilmer and Chen Sun for helpful discussions. RR thanks Mario Lucic and Neil Houlsby for mentoring this research as part of the AI residency, and thanks Basil Mustafa for encouraging conversations.
\fi

{\small
\bibliographystyle{ieee_fullname}
\bibliography{main}
}

\clearpage

\input{appendix.tex}

\end{document}

%% file: table_vtab.tex
\setlength{\tabcolsep}{4pt}
\begin{table*}[!ht]
  \centering
  \normalsize
  \begin{tabular}{l  l l !{\color{lightgray}\vline} c !{\color{lightgray}\vline} c c c}
  \toprule
  \textsc{Method}                 & \textsc{Dataset}      & \textsc{Signal}                              & \text{VTAB}         & \textsc{Natural}       & \textsc{Specialized}       & \textsc{Structured} \\
  \midrule%
  Transitive Invariance  \cite{DBLP:conf/iccv/0004HG17}    & YouTube 100k             &  Tracklets     & 44.2         & 35.0          & 61.8              & 43.4    \\
  MT  \cite{DBLP:conf/iccv/DoerschZ17}  & \ImNetClean \& SoundNet  &  Tracklets     & 59.2         & 51.9          & 78.9              & 55.8    \\
  Supervised (\resnet)                  & \ImNetClean              & None           & \B{68.5}     & \B{71.3}      & \B{83.0}          & \B{58.9} \\
  Detector backbone                     & \openimages              & None           & 61.6         & 60.0          & 80.4              & 53.5      \\
  \midrule
  VIVI \cite{vivi}       & YT8M    &  None     & 60.9     $^{\dagger}$  & 55.0     $^{\dagger}$ & 79.5      $^{\dagger}$  & 56.7     $^{\dagger}$ \\
  OURS                   & YT8M    & Detector  & \B{64.1} $^{\dagger}$  & \B{59.0} $^{\dagger}$ & \B{81.6}  $^{\dagger}$  & \B{59.8} $^{\dagger}$ \\
  \midrule 
  \midrule
  Boxes and labels at random   & YT8M        & None                       & 60.3	    & 55.2	        & 78.0	            & 55.0     \\
  Boxes at random coordinates  & YT8M        & Detector                   & 63.4	    & 57.5	        & 81.1              & 59.7     \\
  Distilling from \ImNetClean  & YT8M        & Classifier                 & 63.1        & 59.6          & \B{81.6}          & 57.0     \\
  Also predict cross entropy   & YT8M        & Detector                   & \B{64.9}    & \B{60.5}      & 81.3              & \B{60.5} \\
  \bottomrule
  \end{tabular}
  \caption{Evaluation on the Visual Task Adaptation Benchmark. Each number indicates the average classification accuracy over all data sets in the corresponding category. ${\dagger}$ indicates a statistical significant difference between the baseline~\cite{vivi} and the proposed method. The last 4 methods show results of 4 ablation studies which investigate the benefits of the corresponding training signals.}
  \label{tab:vtab}
\end{table*}
\setlength{\tabcolsep}{1.4pt}

%% file: table_robustness.tex
\newcommand\rot{90}

\begin{table*}[t]
    \addtolength{\tabcolsep}{3pt}  %
    \centering
     \begin{tabular}{l l c c c c c c c c c !{\color{lightgray}\vline} c c} 
     \toprule 
\textsc{Model} & \textsc{Method} &  \rotatebox[origin=l]{\rot}{  \ImNetClean~\cite{DBLP:conf/cvpr/DengDSLL009}  }	\ & 
                    \rotatebox[origin=l]{\rot}{  \ImNetA ~\cite{DBLP:journals/corr/abs-1907-07174}	} \ & 
                    \rotatebox[origin=l]{\rot}{  \ImNetC ~\cite{DBLP:journals/corr/abs-1807-01697}	} \ & 
                    \rotatebox[origin=l]{\rot}{  \ImNetVV ~\cite{DBLP:conf/icml/RechtRSS19}	} \ & 
                    \rotatebox[origin=l]{\rot}{  \ObjetNet~\cite{DBLP:conf/nips/BarbuMALWGTK19}} \ & 
                    \rotatebox[origin=l]{\rot}{  \ImNetVid~\cite{DBLP:journals/corr/abs-1906-02168}} \ & 
                    \rotatebox[origin=l]{\rot}{  \ImNetVidW~\cite{DBLP:journals/corr/abs-1906-02168}} \ &
                    \rotatebox[origin=l]{\rot}{  \YTBBSHORT~\cite{DBLP:journals/corr/abs-1906-02168}    } \ & 
                    \rotatebox[origin=l]{\rot}{  \YTBBWSHORT~\cite{DBLP:journals/corr/abs-1906-02168}	} \ & 
                    \rotatebox[origin=l]{\rot}{ $\Delta$ \ImNetVid } \ &
                    \rotatebox[origin=l]{\rot}{ $\Delta$ \YTBBSHORT } \\
\midrule %
 & & $\uparrow$ & $\uparrow$ & $\uparrow$ & $\uparrow$ & $\uparrow$ & $\uparrow$ & $\uparrow$ & $\uparrow$ & $\uparrow$ & $\downarrow$ & $\downarrow$ \\
\midrule %
VIVI   & Fine tuning       &    62.6  &    0.5  &    6.8  &    51.1  &    16.2  &    57.9  & 36.5     &    58.0  &    39.9  &    21.4  & \B{18.1}   \\
OURS   & Fine tuning       & \B{65.2} & \B{0.6} & \B{9.5} & \B{53.4} & \B{18.4} & \B{61.7} & \B{43.4} & \B{60.8} & \B{42.3} & \B{18.3} &    18.6   \\
\midrule  %
VIVI   & Cotraining      &    73.1  &    1.1  &    24.3  &    59.8  &    20.9  &    58.7  &    41.7  &    49.4  &    35.4  &    17.0  &    14.0   \\
OURS   & Cotraining      & \B{73.3} & \B{1.2} & \B{24.4} & \B{60.8} & \B{21.0} & \B{59.7} & \B{44.0} & \B{50.0} & \B{37.6} & \B{15.7} & \B{12.4}   \\
\bottomrule
     \end{tabular}%
    \caption{\label{tab:robustness_eval} Accuracy on robustness datasets from literature. These datasets are typically a perturbed version of ImageNet-like images and videos. Each dataset indicates a specific aspect of robustness. Higher accuracy corresponds to better robustness. Lower $\Delta$ corresponds to better robustness. \textit{Cotraining} refers to a model trained on \YTeightM and \ImNetClean.}
\end{table*}

%% file: appendix.tex
\appendix 

\section{Statistics on the annotated \YTeightM}
This section shows statistics on the \YTeightM, annotated with the object detector~\cite{tfhub}. We annotate each frame of \YTeightM with the object detector and store the five objects with highest detection scores. Our method relies on objects recurring multiple times in a video. The method works better when objects occur multiple times in the selected frames. Therefore, \Cref{tab:object_recurrence} displays statistics for objects that occur in most videos. For each object, we count how often the object recurs in the \num{32} frames sampled with the strategy from \cite{vivi}. For example, in \num{49} percent of videos, an object with class \textit{Footwear} occurs. Each of those videos has, on average, \num{15} instances of the \textit{Footwear} class.

\begin{figure}[h]
    \centering
    \includegraphics[width=\linewidth]{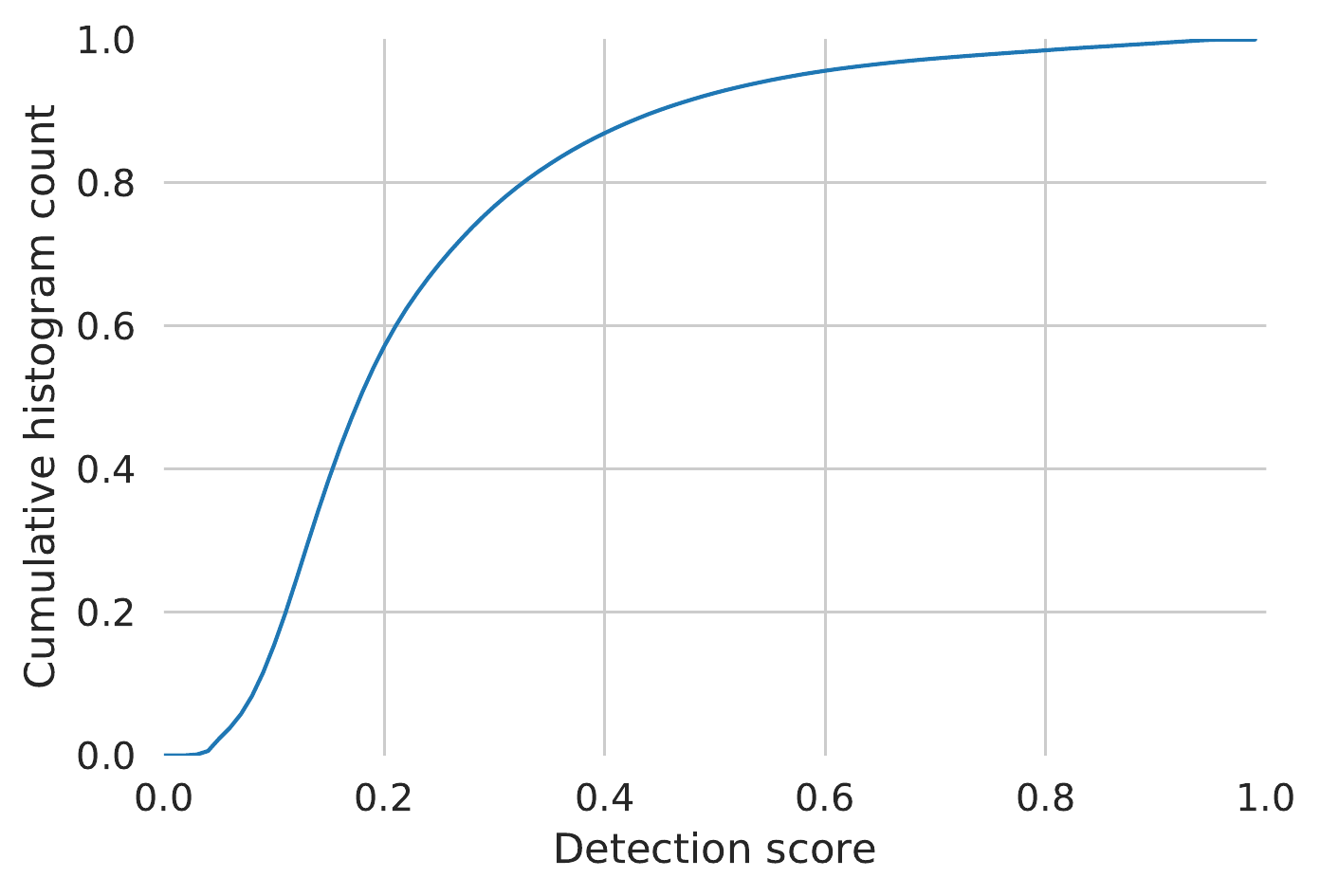}
    \caption{Cumulative histogram of the detection scores from the object detector. Histogram measured on the videos from \YTeightM, annotated with our detector~\cite{tfhub}. In our experiments, we exclude boxes with scores below \num{0.05}, which applies to approximately 3 percent of the objects.}
    \label{fig:hist_det_score}
\end{figure}

We discard objects with a low detection score. \Cref{fig:hist_det_score} shows the fraction of boxes below a certain threshold. All methods in this work use a threshold of \num{0.05}, which discards about 3 percent of the objects. We experimented with higher thresholds, but this resulted in worse VTAB scores.

\begin{table}[h]
    \centering
     \begin{tabular}{l l l} 
     \toprule 
     \textsc{Label name}        & Videos (\%)  & Recurrence  \\
     \midrule
\textsc{ Street light         } & \num{   14 } & \num{ 13.1 } \\
\textsc{ Flower               } & \num{   15 } & \num{  9.6 } \\
\textsc{ Chair                } & \num{   15 } & \num{  7.3 } \\
\textsc{ Land vehicle         } & \num{   15 } & \num{  5.6 } \\
\textsc{ Table                } & \num{   20 } & \num{  7.6 } \\
\textsc{ Toy                  } & \num{   21 } & \num{ 13.0 } \\
\textsc{ Bottle               } & \num{   22 } & \num{  9.7 } \\
\textsc{ Car                  } & \num{   24 } & \num{ 12.0 } \\
\textsc{ Building             } & \num{   28 } & \num{  9.9 } \\
\textsc{ Woman                } & \num{   30 } & \num{ 12.0 } \\
\textsc{ Wheel                } & \num{   30 } & \num{ 13.4 } \\
\textsc{ Poster               } & \num{   37 } & \num{ 11.1 } \\
\textsc{ Window               } & \num{   44 } & \num{ 17.9 } \\
\textsc{ Footwear             } & \num{   49 } & \num{ 15.0 } \\
\textsc{ Tree                 } & \num{   50 } & \num{ 32.9 } \\
\textsc{ Man                  } & \num{   51 } & \num{ 14.8 } \\
\textsc{ Human face           } & \num{   72 } & \num{ 20.5 } \\
\textsc{ Clothing             } & \num{   84 } & \num{ 24.1 } \\
\textsc{ Person               } & \num{   86 } & \num{ 30.0 } \\
     \bottomrule
     \end{tabular}
    \caption{\label{tab:object_recurrence} Recurrence of objects within the 32 frames sampled for learning from one video. For example, on average, \num{86}\% of the videos contain an object labeled \textsc{Person}. In each video where a \textsc{Person} occurs, the detector annotated an average of \num{30} instances. We show averages over ten thousand videos that we randomly sampled from the training set.}
\end{table}

\begin{figure}[h]
    \centering
    \includegraphics[width=\linewidth]{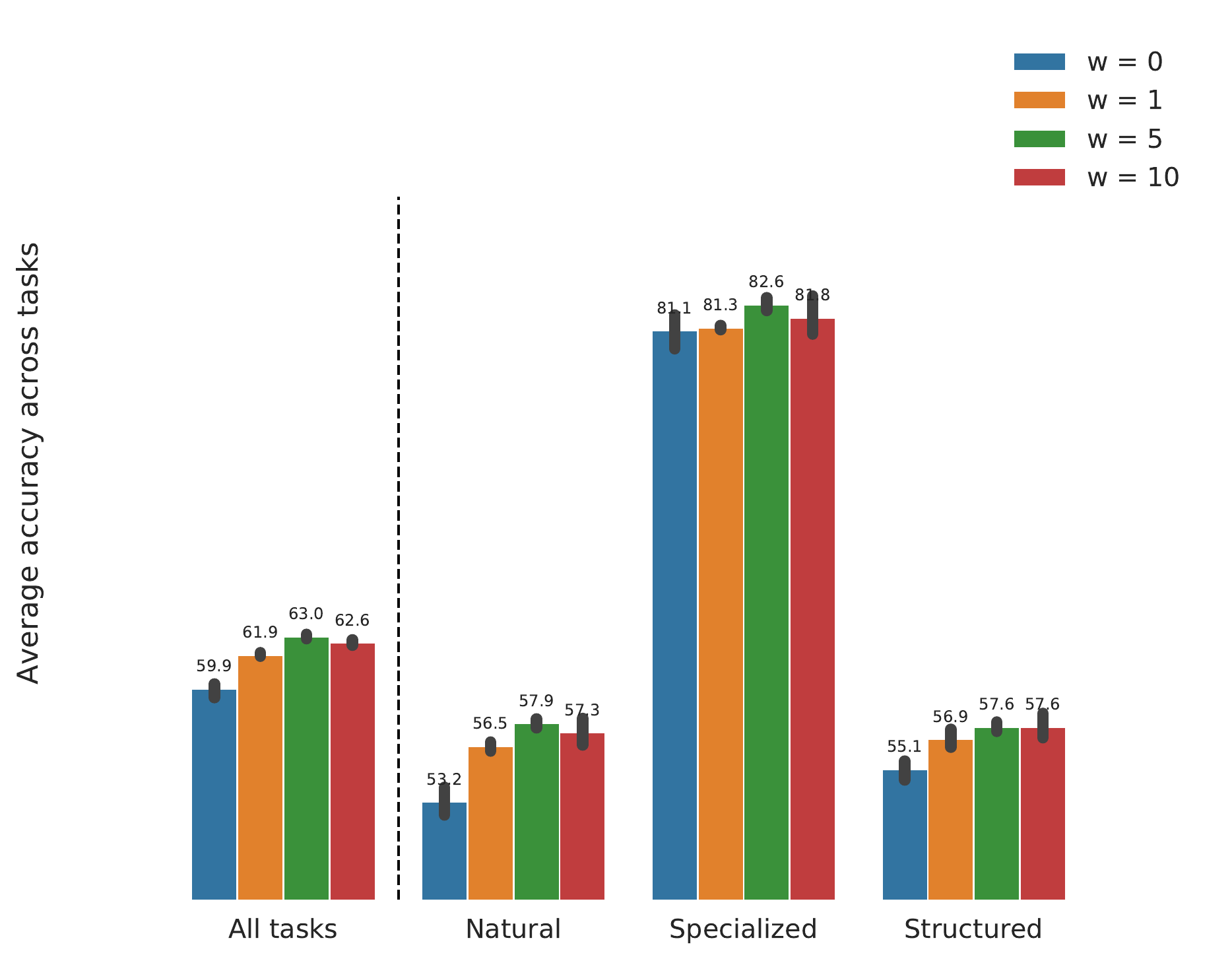}
    \caption{VTAB scores on respective validation sets when changing the weight for the object-level loss. The optimum accuracy occurs at \num{5}, which is the value we use in all experiments. The VTAB scores change away from the optimum, but are relatively stable when comparing to baseline (see \Cref{tab:vtab}). The error bars indicate bootstrapped 95\% confidence intervals. \looseness=-1}
    \label{fig:sensitivity_weight}
\end{figure}

\section{Sensitivity to hyperparameters}

\begin{figure}[h]
    \centering
    \includegraphics[width=\linewidth]{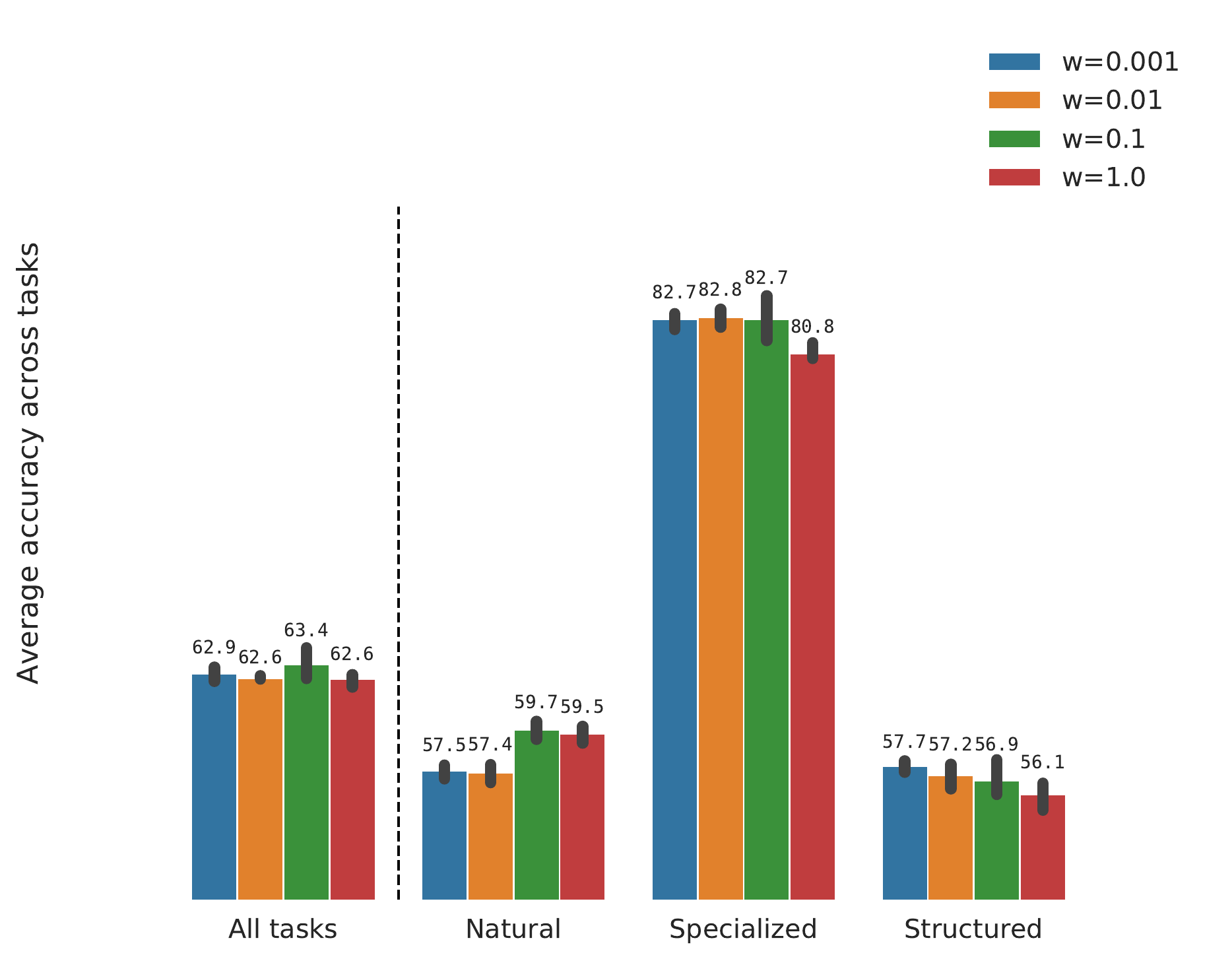}
    \caption{VTAB scores on respective validation sets when changing the weight for the additional supervised loss on the objects. The optimum accuracy occurs at \num{.1}, which is the value we use in the ablation experiment. The error bars indicate bootstrapped 95\% confidence intervals.}
    \label{fig:sensitivity_weight_both}
\end{figure}

\begin{figure}[h]
    \centering
    \includegraphics[width=\linewidth]{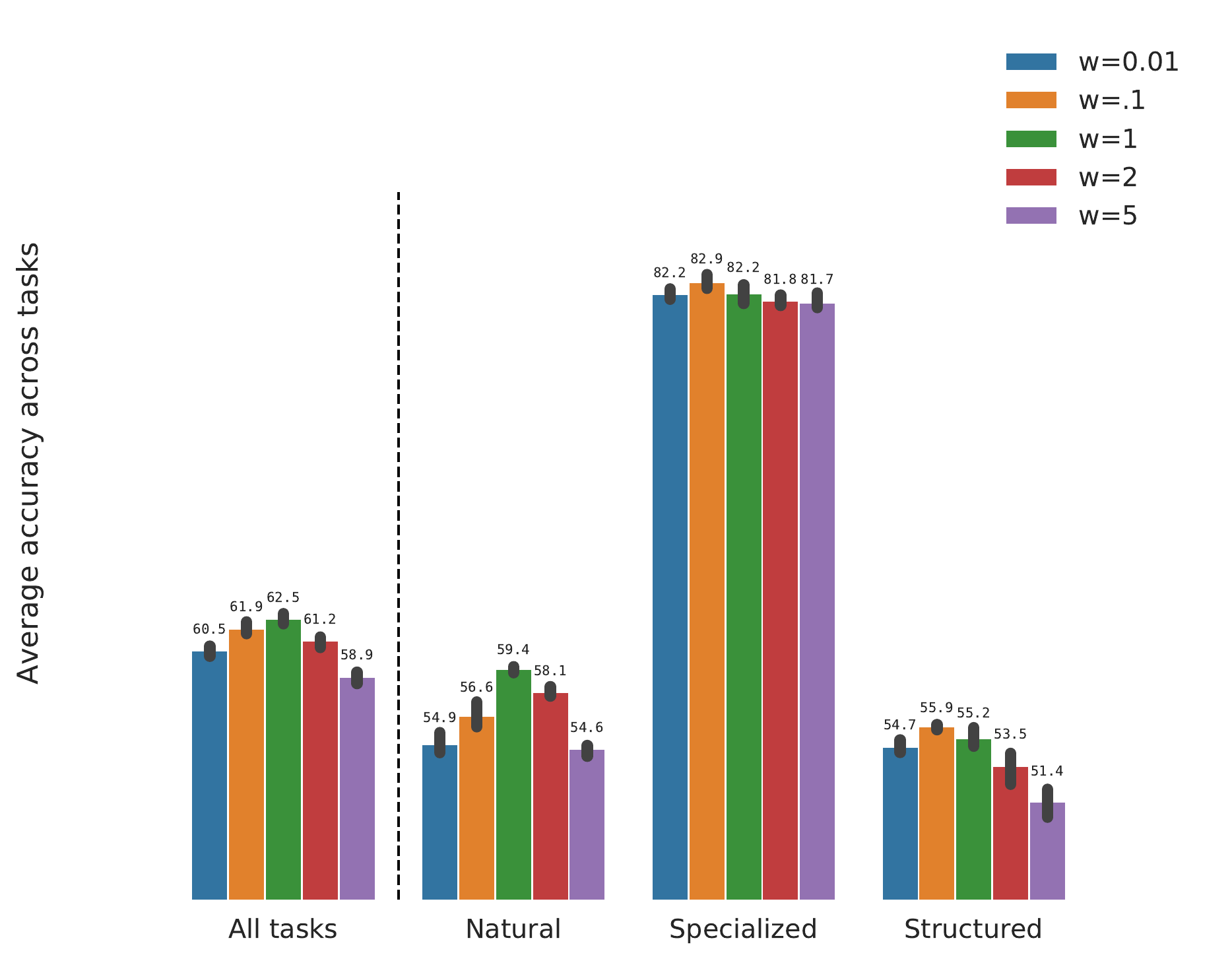}
    \caption{VTAB scores on respective validation sets when changing the weight for cross entropy loss on the soft labels. This corresponds to row  \textit{Distilling from \ImNetClean} reported in \Cref{tab:vtab}. The optimum accuracy occurs at \num{1.0}, which is the value we use for the experiment . The error bars indicate bootstrapped 95\% confidence intervals. \looseness=-1}
    \label{fig:sensitivity_weight_imnet_logits}
\end{figure}

Our experiments use three important hyperparameters. We used the validations sets from the VTAB benchmark to set the hyperparameters. This section shows the sweeps we make so one can judge the sensitivity for each hyperparameter. \Cref{fig:sensitivity_weight} shows the search for hyperparameter $\omega$ from \Cref{eqn:losses_combi}. \Cref{fig:sensitivity_weight_both} shows the search for a positive coefficient to include the cross entropy loss in the experiment for \Cref{tab:vtab}, row \textit{Also predict cross entropy}. \Cref{fig:sensitivity_weight_imnet_logits} shows the search for a positive coefficient for the cross entropy loss when learning from the soft labels from \ImNetClean for \Cref{tab:vtab}, row \textit{Distilling from \ImNetClean}.

\begin{table*}[h!]
    \centering
    \small
    \resizebox{\textwidth}{!}{%
    \input{table_appendix_large.tex}    }
    \caption{VTAB accuracies for each method and dataset considered in our work. Each number represents the accuracy after transferring the model learned with the method to the specific dataset. Each dataset has only 1000 labeled samples. We follow the transfer protocol from \cite{vtabimplementation}}
    \label{tab:large_table}
\end{table*}

%% file: table_appendix_large.tex
\fontsize{7pt}{7pt}\selectfont
\newcolumntype{C}{>{\centering\arraybackslash}X}
\setlength{\tabcolsep}{0pt}
\setlength{\extrarowheight}{5pt}
\renewcommand{\arraystretch}{0.80}
\begin{tabularx}{\linewidth}{C l | CCCC !{\color{lightgray}\vline} CCCCCCC !{\color{lightgray}\vline} CCCC !{\color{lightgray}\vline} CCCCCCCC}
\toprule
&
Method                                         \ & 
\rotatebox[origin=c]{90}{All tasks           } \ & 
\rotatebox[origin=c]{90}{Natural             } \ & 
\rotatebox[origin=c]{90}{Specialized         } \ & 
\rotatebox[origin=c]{90}{Structured          } \ & 
\rotatebox[origin=c]{90}{Caltech101          } \ & 
\rotatebox[origin=c]{90}{Cifar-100           } \ & 
\rotatebox[origin=c]{90}{DTD                 } \ & 
\rotatebox[origin=c]{90}{Flowers102          } \ & 
\rotatebox[origin=c]{90}{Pets                } \ & 
\rotatebox[origin=c]{90}{SUN397              } \ & 
\rotatebox[origin=c]{90}{SVHN                } \ & 
\rotatebox[origin=c]{90}{Retinopathy         } \ & 
\rotatebox[origin=c]{90}{EuroSAT             } \ & 
\rotatebox[origin=c]{90}{Resics45            } \ & 
\rotatebox[origin=c]{90}{Camelyon            } \ & 
\rotatebox[origin=c]{90}{CLEVR-dist          } \ & 
\rotatebox[origin=c]{90}{CLEVR-count         } \ & 
\rotatebox[origin=c]{90}{DMLab               } \ & 
\rotatebox[origin=c]{90}{dSPRITES-orient     } \ & 
\rotatebox[origin=c]{90}{dSPRITES-pos        } \ & 
\rotatebox[origin=c]{90}{sNORB-azimuth       } \ & 
\rotatebox[origin=c]{90}{sNORB-elevation     } \ & 
\rotatebox[origin=c]{90}{KITTI               } \\
\midrule
& \resnet from scratch
& 42.1 & 26.9 & 65.8 & 43.6 & 37.7 & 11.0 & 23.0 & 40.2 & 13.3 &  3.9 & 59.3 & 63.1 & 84.8 & 41.6 & 73.5 & 54.8 & 38.5 & 35.8 & 37.3 & 87.9 & 20.9 & 36.9 & 36.9 \\
& Transitive Invariance~\cite{DBLP:conf/iccv/0004HG17}
& 44.2 & 35.0 & 61.8 & 43.3 & 54.9 &  7.1 & 38.3 & 28.2 & 32.3 &  7.4 & 77.0 & 63.1 & 84.1 & 50.0 & 50.0 & 61.7 & 12.7 & 35.0 & 59.3 & 86.1 & 21.1 & 29.2 & 41.6 \\
& MS~\cite{DBLP:conf/iccv/DoerschZ17}                    
& 47.2 & 33.4 & 68.4 & 47.9 & 52.3 & 12.7 & 37.3 & 32.6 & 15.8 &  6.8 & 81.8 & 57.3 & 89.7 & 49.7 & 76.8 & 55.7 & 43.2 & 38.4 & 46.4 & 81.2 & 34.8 & 35.1 & 48.4 \\
& MT~\cite{DBLP:conf/iccv/DoerschZ17}                    
& 59.2 & 51.9 & 78.9 & 55.8 & 76.2 & 26.2 & 49.3 & 63.5 & 48.5 & 10.6 & 89.1 & 71.7 & 93.3 & 70.2 & 80.3 & 62.1 & 55.6 & 44.3 & 43.2 & 86.6 & 39.1 & 38.9 & 76.3 \\
& MobileNetV2            
& 65.9 & 69.5 & 81.9 & 54.8 & 88.5 & 45.4 & 59.1 & 87.3 & 86.7 & 32.0 & 87.3 & 71.1 & 94.3 & 80.5 & 81.6 & 55.8 & 44.8 & 46.6 & 51.6 & 90.0 & 37.5 & 38.7 & 73.4 \\
& \ImNetClean supervised         
& 68.5 & 71.3 & 83.0 & 58.9 & 84.7 & 60.0 & 68.2 & 87.3 & 90.3 & 36.1 & 72.2 & 75.4 & 95.2 & 81.4 & 80.0 & 57.7 & 73.9 & 45.6 & 59.7 & 88.5 & 29.1 & 34.2 & 82.3 \\
& \ImNetClean supervised (3x)
& 69.5 & 72.6 & 83.8 & 59.5 & 85.6 & 61.0 & 69.6 & 88.8 & 90.9 & 37.4 & 75.0 & 78.0 & 95.7 & 82.5 & 78.9 & 61.4 & 64.6 & 45.3 & 60.5 & 93.2 & 32.9 & 36.6 & 81.5 \\
& Detector backbone~\cite{tfhub}      
& 61.6 & 60.0 & 80.4 & 53.5 & 84.3 & 38.2 & 48.4 & 77.4 & 58.6 & 25.2 & 88.1 & 70.6 & 94.0 & 73.4 & 83.5 & 58.2 & 42.8 & 47.8 & 46.4 & 73.4 & 39.4 & 42.9 & 77.4 \\
\multirow{-9}{*}{\rotatebox{90}{Comparison}}
& BigBiGAN~\cite{bigbigan}        
& 59.1 & 56.6 & 79.1 & 51.3 & 80.8 & 39.2 & 56.6 & 77.9 & 44.4 & 20.3 & 76.8 & 69.3 & 95.6 & 74.0 & 77.4 & 55.6 & 53.9 & 38.7 & 46.7 & 70.6 & 27.2 & 46.3 & 71.4 \\
\midrule
& VIVI~\cite{vivi} 
& 60.8 & 55.1 & 80.0 & 56.3 & 74.8 & 29.2 & 48.6 & 76.9 & 54.8 & 13.6 & 87.6 & 71.4 & 94.4 & 74.1 & 80.1 & 59.0 & 54.0 & 47.1 & 50.9 & 91.7 & 37.0 & 42.4 & 68.2 \\
& OURS                   
& 64.0 & 58.9 & 81.8 & 59.7 & 81.5 & 35.9 & 51.6 & 76.9 & 60.1 & 17.1 & 89.3 & 72.7 & 94.7 & 76.9 & 82.7 & 62.6 & 61.5 & 50.8 & 53.3 & 92.2 & 41.5 & 39.0 & 76.6 \\
& Rand boxes and labels\hspace{1pt} 
& 60.3 & 55.1 & 80.0 & 54.9 & 75.7 & 28.2 & 49.6 & 76.7 & 53.1 & 14.7 & 88.0 & 71.3 & 93.8 & 74.0 & 80.8 & 60.8 & 55.7 & 34.5 & 50.7 & 94.0 & 37.2 & 37.0 & 69.5 \\
& Rand boxes             
& 63.4 & 57.5 & 81.1 & 59.7 & 79.4 & 31.4 & 51.3 & 77.0 & 58.8 & 16.2 & 88.5 & 72.2 & 94.3 & 74.5 & 83.5 & 61.3 & 60.0 & 48.0 & 52.2 & 95.0 & 40.6 & 42.1 & 78.2 \\
& Distilling from ImNet      
& 63.1 & 59.6 & 81.5 & 56.9 & 81.3 & 35.4 & 58.4 & 75.5 & 54.3 & 24.9 & 87.5 & 73.0 & 95.4 & 75.8 & 82.0 & 61.0 & 50.0 & 47.0 & 50.7 & 89.3 & 36.5 & 41.8 & 79.3 \\
& Include CE loss        
& 64.9 & 60.5 & 81.3 & 60.5 & 83.9 & 38.9 & 55.2 & 76.2 & 59.2 & 20.7 & 89.3 & 71.2 & 94.7 & 77.0 & 82.3 & 63.7 & 65.6 & 50.8 & 52.8 & 94.2 & 34.7 & 40.4 & 81.7 \\
& Distilling detector\hspace{1pt}
& 57.1 & 52.2 & 77.2 & 51.3 & 78.2 & 29.6 & 49.1 & 56.9 & 47.2 & 21.0 & 83.5 & 70.0 & 91.8 & 66.3 & 80.5 & 58.1 & 44.9 & 41.4 & 44.9 & 77.5 & 30.9 & 32.4 & 80.2 \\
\multirow{-8}{*}{\rotatebox{90}{Video only}} 
& Filter half of detections 
& 61.9 & 57.6 & 80.3 & 56.4 & 79.7 & 31.8 & 50.4 & 78.3 & 59.2 & 14.5 & 89.2 & 71.0 & 94.3 & 74.9 & 81.0 & 58.4 & 50.6 & 48.4 & 52.2 & 90.9 & 34.9 & 41.6 & 74.4 \\

\midrule
& VIVI~\cite{vivi}                   
& 69.0 & 70.0 & 83.5 & 60.8 & 87.2 & 51.5 & 64.6 & 88.2 & 85.9 & 32.3 & 79.9 & 72.5 & 95.5 & 81.0 & 84.9 & 61.2 & 74.6 & 44.7 & 61.9 & 90.6 & 29.4 & 43.7 & 80.5 \\
& OURS                   
& 69.5 & 70.7 & 83.2 & 61.5 & 88.0 & 53.1 & 64.9 & 88.1 & 86.3 & 33.5 & 81.0 & 72.2 & 94.5 & 80.5 & 85.6 & 56.5 & 79.3 & 46.6 & 60.5 & 92.7 & 28.6 & 45.2 & 82.3 \\
& VIVI (3x)~\cite{vivi}              
& 70.5 & 72.6 & 83.8 & 62.0 & 88.0 & 54.3 & 69.4 & 89.6 & 87.9 & 34.6 & 84.2 & 72.9 & 95.3 & 82.3 & 84.9 & 58.3 & 74.5 & 46.3 & 67.8 & 92.1 & 33.1 & 44.1 & 80.2 \\
\multirow{-4}{*}{\rotatebox{90}{Cotraining}}
& OURS (3x)              
& 70.8 & 72.2 & 83.4 & 63.4 & 87.1 & 55.3 & 67.8 & 90.0 & 87.7 & 35.6 & 81.6 & 72.0 & 95.0 & 82.4 & 84.1 & 63.2 & 80.5 & 47.3 & 66.0 & 87.8 & 33.9 & 46.0 & 82.6 \\
\bottomrule
\end{tabularx}